\documentclass[11pt]{article}

\usepackage[preprint]{acl}

\usepackage{newtxtext} % 更好的 Times 文本字体
\usepackage{newtxmath} % 更好的 Times 数学字体
\usepackage{latexsym}

\usepackage[T1]{fontenc}
\usepackage[utf8]{inputenc}

\usepackage{microtype}

\usepackage{inconsolata}
\usepackage{graphicx}
\usepackage{booktabs}
\usepackage{enumitem}
\usepackage{wrapfig}
\usepackage{algorithm}
\usepackage{algpseudocode}
\usepackage{graphicx}
\usepackage{caption}
\usepackage{subcaption}
\usepackage[misc]{ifsym}

\usepackage{microtype}
\usepackage{amsmath}
\usepackage{colortbl}
\usepackage[utf8]{inputenc}
\usepackage[T1]{fontenc}
\definecolor{lightgray}{rgb}{0.9,0.9,0.9}
\usepackage{caption}
\usepackage{subcaption}
\usepackage{setspace}
\usepackage{url}
\usepackage{multirow}
\usepackage{tabularx}
\usepackage{blindtext}
\usepackage{pgfplots}
\usepackage{tikz}
\usetikzlibrary{er,positioning,bayesnet}
\usepackage{makecell}
\usepackage{tipa}
\usepackage{siunitx}
\usepackage{nicefrac}
\usepackage{listings}
\usepackage[raster,skins, most]{tcolorbox} %
\usepackage{xltabular}
\usepackage{adjustbox}
\usepackage{xurl}
\usepackage{rotating}
\usepackage[normalem]{ulem}

\usepackage{amsmath}   % 提供 equation 环境、对齐、\text 等
\usepackage{amsfonts}  % 提供 \mathcal, \mathbb 等字体
\usepackage{bm}        % 粗体数学符号（如果需要）
\usepackage{bbm}

\usepackage{fontawesome}

\title{AutoForge: Automated Environment Synthesis for Agentic Reinforcement Learning}

\author{Shihao Cai\footnotemark[1], Runnan Fang\footnotemark[1], Jialong Wu, Baixuan Li, Xinyu Wang\footnotemark[2], Yong Jiang, \\ \textbf{Liangcai Su, Liwen Zhang, Wenbiao Yin, Zhen Zhang, Fuli Feng, Pengjun Xie, Xiaobin Wang\footnotemark[2]}
\\ Tongyi Lab, Alibaba Group}

\begin{document}
\maketitle

\newcommand\blfootnote[1]{%
\begingroup
\renewcommand\thefootnote{}\footnote{#1}%
\addtocounter{footnote}{-1}%
\endgroup
}

\blfootnote{*These authors contributed equally to this work.}
\blfootnote{\dag Corresponding author.}

\begin{abstract}
Conducting reinforcement learning (RL) in simulated environments offers a cost‑effective and highly scalable way to enhance language-based agents.
However, previous work has been limited to semi‑automated environment synthesis or tasks lacking sufficient difficulty, offering little breadth or depth.
In addition, the instability of simulated users integrated into these environments, along with the heterogeneity across simulated environments, poses further challenges for agentic RL.
In this work, we propose: (1) a unified pipeline for automated and scalable synthesis of simulated environments associated with high-difficulty but easily verifiable tasks; and (2) an environment level RL algorithm that not only effectively mitigates user instability but also performs advantage estimation at the environment level, thereby improving training efficiency and stability.
Comprehensive evaluations on agentic benchmarks, including 
$\tau$-bench, $\tau^2$-Bench, and VitaBench, validate the effectiveness of our proposed method. 
Further in‑depth analyses underscore its out-of-domain generalization.
\end{abstract}
\section{Introduction}
As large language models (LLMs) have demonstrated powerful tool‑use capabilities~\cite{shen2024llm,team2025tongyi,wang2025klear} and robust multi‑turn conversational skills~\cite{yi2024survey},  the research community has increasingly been building language-based agents to automate task solving~\cite{wang2024survey, su2025scaling}.
However, current LLM‑based agents still struggle with complex real‑world tasks that require extensive interaction with environments and users~\cite{he2025vitabench}.
Reinforcement learning (RL) in real‑world environments appears to be an effective way to enhance agent capabilities, but it suffers from high costs and poor scalability~\cite{zhao2025mua}.
Therefore, synthesizing simulated environments for agentic RL has emerged as a popular approach~\cite{qian2025userrl}.

While some recent efforts have explored training agents with mock environments and simulated users~\cite{xi2025agentgym,ftrl,codegym}, they fall short in three key aspects:  
(1) Prior work~\cite{agentscaler,qian2025userrl} has been limited to semi‑automated environment synthesis or tasks that are not sufficiently challenging, falling short in both the breadth and depth required for effective agent training;
(2) Some studies~\cite{zhao2025mua} introduce simulated LLM-based users to interact with agents, yet neglect the instability of such users;
%(3) Existing approaches do not fully exploit multiple synthesized environments for RL~\cite{qian2025userrl}, resulting in suboptimal efficiency and stability.  
(3) Existing approaches~\cite{qian2025userrl} view multi-environment RL training from a single-environment perspective, leading to suboptimal efficiency and stability.

We thus argue that a unified pipeline capable of automatically generating mock environments and complex tasks is better suited to simultaneously broadening the scope of agent training and deepening its complexity.
In addition, during the interaction between the agent and the simulated user, we emphasize the importance of real-time monitoring of the simulated user’s accuracy, so as to ensure that the feedback received by the agent is reasonable. 
Furthermore, we highlight treating agentic RL at the multi‑environment level to improve training efficiency and stability.

To this end, we introduce \textbf{AutoForge}, a novel framework that integrates an automated pipeline for synthesizing environments and challenging tasks, together with an Environment-level Relative Policy Optimization (ERPO) algorithm.
Specifically, our synthesis pipeline starts from tool description documentation, enabling the automated construction of a database to store environment states and the generation of tool implementations in Python. 
A dependency graph of the tools is then constructed, upon which random walks yield diverse tool sequences. 
These sequences are merged and augmented with reasoning nodes and edges to form a complex directed acyclic graph (DAG), which in turn serves as the blueprint for producing tasks.
To address instability in the simulated user, we employ an LLM-as-judge mechanism during the RL rollout phase to identify and mask trajectories where task failures are due to simulated user errors, thereby promoting fairness and reducing bias in advantage estimation.
Furthermore,  we extend the native GRPO’s group-level advantage estimation~\cite{shao2024deepseekmath} to an \textbf{\textit{environment-level}} formulation, which effectively mitigates the impact of outlier samples on the standard deviation. This enhancement improves the accuracy of advantage estimation and contributes to greater training stability.

To validate the effectiveness of the proposed framework, we conduct comprehensive experiments on three popular agentic benchmarks: $\tau$-bench, $\tau^2$-Bench and VitaBench.
Experimental findings indicate that AutoForge markedly enhances the agentic capacity to solve tasks after engaging in extended multi-turn interactions with the environment and the user.
Further in-depth analysis reveals that AutoForge exhibits strong out-of-domain generalization capability as well as robust training stability.
In conclusion, our main contributions are summarized as follows:
\begin{itemize}[topsep=0pt,itemsep=0pt,parsep=0pt,leftmargin=*]
    \item We present a unified pipeline capable of automatically building mock environments and producing high-difficulty tasks, requiring only a textual description of the tool.
    \item We introduce ERPO, an agentic RL algorithm that effectively mitigates the instability caused by simulated users while enabling more robust environment-level advantage estimation.
    \item Comprehensive experimental results confirm both the effectiveness and strong generalization of AutoForge, underscoring the promise and importance of leveraging mock environments in agentic reinforcement learning.
    
\end{itemize}
\section{Related Work}

%\subsection{Agentic Reinforcement Learning}
\paragraph{Agentic Reinforcement Learning.}
For an agent learning within a given environment, the optimization objective is to maximize cumulative reward through a sequence of interactions with the environment.
A growing body of recent work studies agents that extend LLMs with external tools.
% channels: search engines, Python interpreters, or MCP-based tools, and learn to wield them through reinforcement learning. 
Search agents~\cite{li2025webthinker,wu2025webdancer,li2025websailor,tao2025webshaper} interleave reasoning with real-time evidence retrieval to solve challenging open-domain questions. 
In the code-generation arena, CodeRL~\cite{le2022coderl} and DataMind~\cite{qiao2025scaling} treat the interpreter as an executable action space and optimize code-writing policies via reward signals received from execution feedback. 
For general tool manipulation, ToolRL~\cite{qian2025toolrl} and ToolN1~\cite{chen2025toolexpander} dispense with dense instruction tuning and instead train LLMs to select and invoke APIs by directly maximizing task success.
% , demonstrating that ``reward is all tool learning need''. 
% Despite their diverse domains, these systems share a unified paradigm: equip an LLM with modular tools, formulate tool calls as atomic actions, and refine the resulting interactive policy with off-policy or on-policy RL. 
Empirical results show that environment-grounded optimization extends the reasoning horizon of language models but raises challenges in sample efficiency, credit assignment, and training stability.
In our paper, we accelerate RL training by synthesizing a sandbox environment to get efficient feedback.
Furthermore, we improve credit assignment and training stability by masking user-induced hallucination errors and estimating advantages at the environment level.
%\subsection{Environment Emulation}
\paragraph{Environment Emulation.}
In contrast to traditional reinforcement learning, agentic reinforcement learning requires the agent to interact with the environment.
When agents interact with search engines~\cite{wu2025webdancer,li2025websailor}, MCP services~\cite{xu2025toucan,ren2025gtm}, or other real-world tools, they spend most of their wall-clock time waiting.
% : every action pauses until the external world returns a reward, a state, or simply an HTTP response. 
% These idle \textit{bubble} minutes dominate the entire experiment.
Idle \textit{bubble} minutes account for most of the experimental time.
% In practice, idle \textit{bubble} minutes occupy a large portion of the total experimental time.
To address this latency issue, numerous works have explored environment simulation to get efficient feedback. 
On one hand, approaches such as Toolbench\cite{qin2023toolllm}, ToolSandbox\cite{guo2024stabletoolbench}, ZeroSearch\cite{sun2025zerosearch}, train a model-based environment simulator. 
However, due to inherent hallucinations in LLMs, the same agent action may yield inconsistent feedback, and the diversity of the trained model’s outputs introduces instability, limiting the reliability of such model-based simulators.
On the other hand, works like $\tau$-bench\cite{taubench,tau2}, construct local, executable environments, e.g., by converting tasks into locally runnable code or databases. Yet these approaches heavily rely on manual annotation, and automating the construction\cite{shi2025taskcraft} of complex, diverse tasks remains highly challenging.
In this paper, we propose a method for automatically constructing RL training environments and tasks for agents. 
% Via task merging, reasoning node insertion, and user obfuscation—to automatically enhance task complexity. 
% This yields a highly efficient and stable training environment, paired with appropriately challenging tasks, thereby better supporting subsequent RL training.

\begin{figure*}[htbp]
\centering
\includegraphics[width=0.9\textwidth]{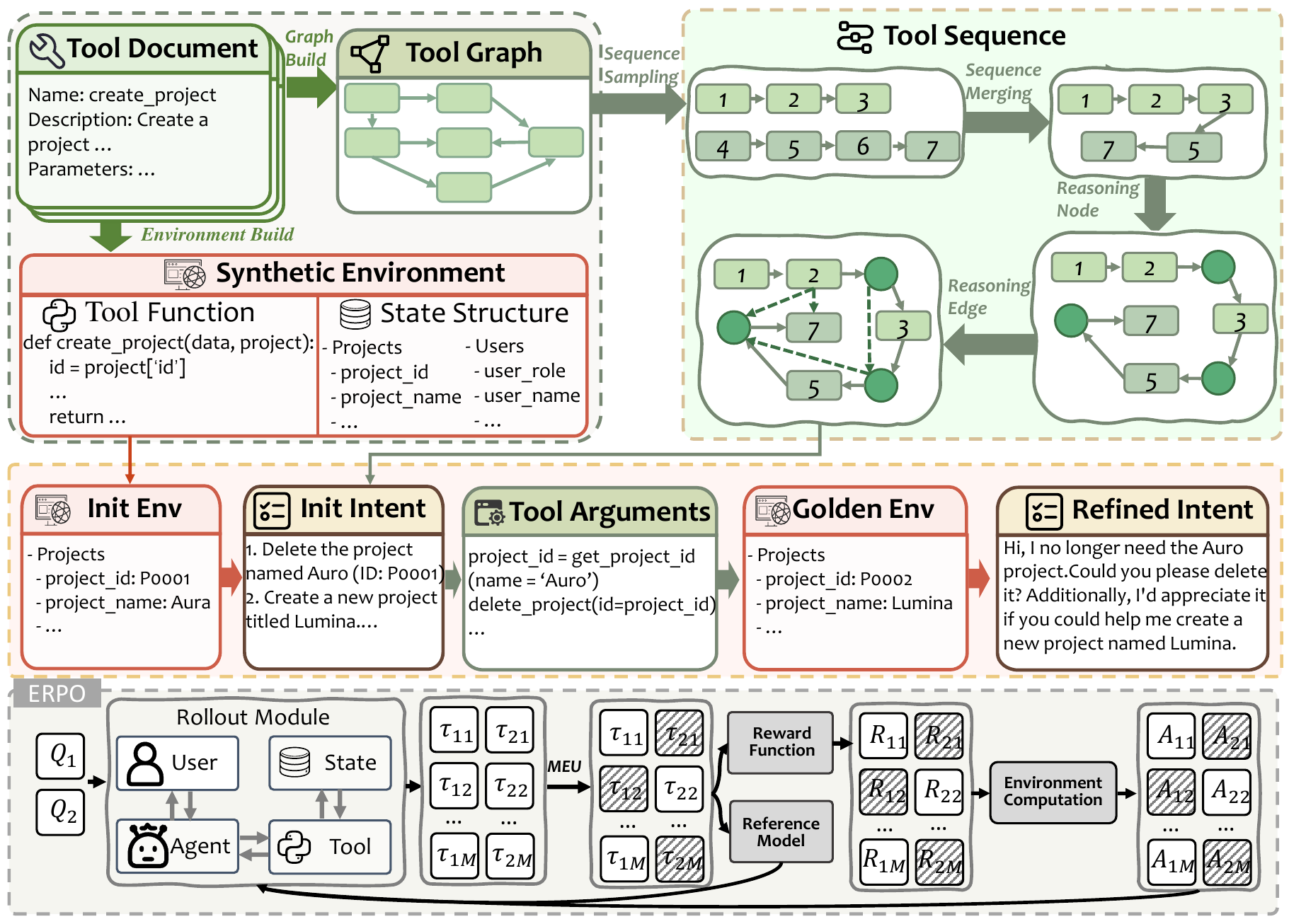}
    \caption{The AutoForge framework consists of a unified pipeline for scalable synthesis of simulated environments and high‑difficulty, easily verifiable tasks, and ERPO algorithm for multi‑environment agentic RL.
}
\label{fig:pipeline}
\end{figure*}
\section{Method}

In this section, we first introduce the verifiable interactive‑environment synthesis pipeline, which consists of three key steps: 
(1) environment synthesis (Section~\S\ref{sec:env_synthesis}), 
(2) tool‑sequence generation (Section~\S\ref{sec:tool_seq_gen}), and 
(3) task generation (Section~\S\ref{sec:task_gen}).
We then present our environment-level agentic RL algorithm (Section~\S\ref{sec:erpo}).

\subsection{Environment Synthesis}
\label{sec:env_synthesis}
Formally, we define a synthetic environment $\mathcal{E}$ as a tuple comprising the state $S$ and an operation function set $\mathcal{F}$, such that $\mathcal{E} = (S, \mathcal{F})$.
For scalable and automated synthesis of environments, we leverage the gathered tool‑description documentation to automatically construct the state $S$ and the corresponding function set $\mathcal{F}$ applicable to the state.

\textbf{State Structure Generation}.
To ensure both the determinism and the flexible extensibility of the state, we record it in a database, 
which can be formally represented as a list of key--value pairs:
$S = [(K_1, V_1), (K_2, V_2), \dots, (K_n, V_n)],$
where $K_i$ denotes the $i$-th attribute name (e.g., \texttt{"project\_id"}), and $V_i$ denotes its corresponding value (e.g., \texttt{"P0001"}).
In this stage, we begin by prompting a LLM to produce the state structure, namely all attribute names $K_i$, 
while leaving the generation of specific attribute values $V_i$ to Section~\S\ref{sec:env_init} in order to better ensure the consistency between tasks and environments.

\textbf{Function Set Generation}.
It is important to note that a function executable on a given state requires the state structure rather than the specific state values. 
Therefore, once the state structure is obtained, we leverage the tool‑description documents together with the state structure to prompt an LLM for generating the corresponding Python code. 
The Python code not only conforms to the state structure but also offers extremely low execution cost, high concurrency, and strong stability, which fully satisfy the requirements of RL training.

\subsection{Tool‑Sequence Generation}
\label{sec:tool_seq_gen}
After building the synthetic simulation environment, we further construct challenging tasks that align with the environment to facilitate RL training for the agent.
We posit that the difficulty of a task is determined by the specific sequence of tools needed to accomplish it. Therefore, our first step is to design tool sequences of high complexity.

\textbf{Sequence Sampling}. Motivated by AgentScaler~\cite{agentscaler}, we begin by representing all available tools as nodes in a directed graph, derived from their description documents. 
A directed edge between two tools is established when an LLM judges that the output of one tool is likely to be a valid input for another.
Subsequently, we perform random walks on the graph to obtain thousands of tool sequences.
However, tool sequences obtained through simple sampling may not reflect complex tasks, such as those involving multiple requirements or intricate reasoning. 
We therefore enhance their complexity to enable the generation of more sophisticated tasks in later stages.

\textbf{Tool Sequence Merging.}
Formally, the $i$-th tool sequence can be denoted as 
$\mathcal{T}_i = [t_{i_{1}}, t_{i_{2}}, \dots, t_{i_{|\mathcal{T}_i|}}]$. 
To obtain a tool sequence that encompasses multiple requirements, 
we can merge two existing sequences $\mathcal{T}_i$ and $\mathcal{T}_j$ 
to produce a new sequence $\mathcal{T}_{k}'$.
However, directly concatenating $\mathcal{T}_i$ and $\mathcal{T}_j$ may introduce redundant tools. 
For instance, $\mathcal{T}_i$ may contain the tool \texttt{find\_id\_by\_name}, 
while $\mathcal{T}_j$ includes \texttt{find\_id\_by\_email}, 
resulting in a merged tool sequence that repeatedly retrieves the same \texttt{id} information. 
 To address this, we prompt an LLM to remove redundant tools, yielding a more coherent merged sequence.

\textbf{Reasoning Node Integration.}
Subsequently, to ensure that a tool sequence captures complex reasoning, we employ an LLM to insert reasoning nodes into the sequence. 
A reasoning node is defined as a node that performs inference on the outputs of preceding nodes 
to derive higher-level information. 
For example, if a preceding node outputs the prices of all items, 
a reasoning node may compute the total price of these items. 
After this step, we obtain tool sequences enriched with higher-order reasoning, formally denoted as
$
\mathcal{T}_k'' = [t_{k_1}, \dots, r_{k_1}, \dots, t_{k_i}, r_{k_j}, \dots, t_{k_{|\mathcal{T}_k''|}}],
$
where $r_{k_j}$ denotes the $j$-th reasoning node in $\mathcal{T}_k''$.

\textbf{Reasoning Edge Integration.}
Finally, to incorporate explicit dependency information among tools, we utilize an LLM to augment the tool sequence 
$\mathcal{T}_k''$ with a set of directed edges, referred to as reasoning edges 
$E_k$. 
This results in a directed acyclic graph (DAG) formally represented as
$
G_k = (\mathcal{T}_k'', E_k).
$
A reasoning edge is defined as a directed connection where 
the output of the parent node is used, through a reasoning process, 
to generate the input parameters for the child node. 

\subsection{Task Generation}
\label{sec:task_gen}
Using $G_k$ as a blueprint, we generate tasks that guide the agent’s interaction with the synthetic environment.

\textbf{Environment Initialization.}
\label{sec:env_init}
As discussed in Section~\S\ref{sec:env_synthesis}, we have already synthesized the environment structure. 
Therefore, we first instantiate the environment structure, 
which means generating a corresponding value $V_i$ for each attribute name $K_i$.
Similar to previous work~\cite{tau2}, we simulate users in the environment and use the user intent to represent the task.
Given the instantiated environment $S_k$ and the graph $G_k$, we prompt an LLM to generate 
the corresponding initial task question (intent) $\tilde Q_k$.

\textbf{Tool Sequence Execution.}
A key question is how to verify whether the agent has successfully completed the task. 
Although we have the gold tool sequence $\mathcal{T}_k''$, 
we choose to evaluate task completion based on the final environment state 
rather than the tool sequence. 
This is because a single task may be solvable through multiple valid sequences of tool calls. 
To obtain the final environment state, we use the initial task $\tilde Q_k$ to fill in the 
tool arguments for each tool name in $G_k$ following its topological order, 
and then execute these tools on the initial environment state $S_k$, 
resulting in the final state $S_k^{*}$.

\textbf{Task Refinement.}
Finally, we use the initial and final environment states, $S_k$ and $S_k^{*}$, to refine the task $\tilde Q_k$. 
A well-crafted task description should contain only the minimal information necessary to solve the task, while remaining linguistically natural. 
This process yields an RL training sample 
$D_k = (Q_k, S_k, S_k^{*}, \mathcal{F})$, where $Q_k$ is the \textbf{refined} version of the task.

\subsection{ERPO}
\label{sec:erpo}
In this section, we leverage the constructed dataset $\mathcal{D}$ to train the agent to continuously interact with the environment and ultimately solve the task.
To better simulate real-world scenarios where the user continuously issues requests 
and the agent interacts with both the user and the environment, 
we introduce a simulated \emph{user agent} into the synthetic environment during the RL stage. 
The user agent, given the task $Q$, continuously provides relevant information. 
Formally, at step $t$, the agent’s action $a_t$ can be one of two options: 
(1) call a tool, or (2) request information from the user agent. 
The corresponding observation $o_t$ is: 
(1) the result returned after tool execution, or 
(2) a textual reply from the user agent.
We extend the vanilla GRPO algorithm~\cite{shao2024deepseekmath} in the following four aspects towards environment-level agentic reinforcement learning.

\textbf{User‑Centered Rollout.}
The rollout phase begins with the user agent generating the first request text $o_0$ based on $Q$. The agent then autonomously decides whether to call a tool or request information from the user agent. The rollout terminates when the user agent determines that all requirements in $Q$ 
have been satisfied. 
Formally, a rollout trajectory can be represented as:
$
\tau = (o_0, a_1, o_1, a_2, o_2, \dots, a_n, o_n),
$
where $a_t$ denotes the action at step $t$ and $o_t$ the corresponding response.
After the rollout, we obtain an environment state $\hat{S}$, which is compared with the golden environment state \(S^{*}\) to compute the reward \(R\).
Therefore, the function for $R$ is as follows:
\begin{equation}
    R =
\begin{cases}
1, & \text{if \(S^{*}\) == $\hat{S}$} \\
\text{0}, & \text{if \(S^{*}\) != $\hat{S}$}
\label{eq:reward_function}
\end{cases}
\end{equation}

\textbf{Interleaved Thinking.}
%We adopt Qwen3~\cite{qwen3} as the training backbone, which discards the thinking content from earlier turns in multi-turn generation to reduce the overall context length. 
We use Qwen3~\cite{qwen3} as the training backbone, which discards previous reasoning content when a new user query is received.
However, we emphasize that for agent tasks requiring multiple rounds of interaction with the user, it is important to retain previous reasoning content, which often includes task analysis and planning.
%We maintain that such intermediate content, typically involving task analysis, reasoning, and planning, plays a crucial role in informing later decisions.
To this end, we modify both the training and inference procedures to preserve all thinking content, thereby enabling \textit{Interleaved Thinking}, as similarly explored in Tongyi DeepResearch~\cite{team2025tongyi}, GPT-OSS~\cite{gptoss} and Minimax-M2~\cite{minimax}.

\textbf{Masking Out Erroneous User Behaviors.}
It is worth noting if during the rollout, the user agent make mistakes, such as returning hallucinated information or omitting information contained in $Q$, which can prevent the agent from completing the task. 
To address this, we employ an LLM to identify trajectories in which the user agent has erred 
and mask them out so that they \textbf{do not} contribute to advantage or loss computation, thereby ensuring the accuracy of advantage estimation. 
The detailed prompt is provided in Appendix~\ref{sec:app_meu}.

For each input \( D \), GRPO~\cite{shao2024deepseekmath} samples a group of trajectories \( \{ \tau_1, \tau_2, \dots, \tau_M \} \) from the old policy model \( \pi_{\text{old}} \).
After applying masking of erroneous user behaviors (\textcolor{red}{MEU}), the RL objective can be expressed as:
\begin{equation}
\small
\begin{aligned}
&J_{\text{ERPO}}(\theta) = 
\mathbb{E}_{D \sim \mathcal{D}, \{\tau_i\}_{i=1}^M \sim \pi_{\pi_{\text{old}}}(\cdot \mid D)} 
\Bigg[ 
\frac{1}{\color{red}{\sum_{i=1}^M \mathbf{1}_{\text{MEU}}(\tau_i)}} \\
&
\textcolor{red}{\sum_{i=1}^M \mathbf{1}_{\text{MEU}}(\tau_i) }\times  
\min \Big( \frac{\pi_{\theta}(\tau_{i} | D)}{\pi_{\text{old}}(\tau_{i} | D)} A_{i}, 
\text{clip} ( \frac{\pi_{\theta}(\tau_{i} | D)}{\pi_{\text{old}}(\tau_{i} | D)},\\&  1 - \epsilon, 1 + \epsilon ) A_i \Big)  - \beta D_{\text{KL}}(\pi_{\theta} \parallel \pi_{\text{old}}) 
\Bigg],
\end{aligned}
\label{eq:grpo}
\end{equation}
where \( \epsilon \) and \( \beta \) are hyperparameters, \(A_i\) is the advantage function, and \(\mathbf{1}_{\mathrm{MEU}}(\tau_i)\) is an indicator that equals \(0\) if the user agent makes an error in trajectory \(\tau_i\), and \(1\) otherwise.

\textbf{Environment-Level Advantage Estimation.}
Formally, a training batch contains multiple environments. Suppose each environment includes \(P\) questions \(\{Q_i\}_{i=1}^P\). For each question \(Q_i\), we sample \(M\) trajectories, and after MEU filtering, obtain \(M_i\) valid trajectories \(\{\tau_{ij}\}_{j=1}^{M_i}\) with the corresponding rewards \(\{R_{ij}\}_{j=1}^{M_i}\).
The original GRPO computes the group-level advantage using all trajectories from the same question, as follows:
\begin{equation}
\begin{aligned}
\small
 A_{ij}^{\text{group}} = \frac{R_{ij} - mean(\{R_{ij}\}_{j=1}^{M_i})}{std(\{R_{ij}\}_{j=1}^{M_i})}.
\label{eq:grpo_adv}
\end{aligned}
\end{equation}

% However, we argue that the group-level standard deviation estimate is susceptible to extreme values, whereas the environment-level estimate is more stable. Therefore, we extend the \textcolor{blue}{group-level advantage to the environment-level advantage} as follows:
We define an \textcolor{blue}{environment-level advantage} by replacing the group-level normalization with performing normalization within each environment. Concretely, the advantage is computed as:
\begin{equation}
\begin{aligned}
\small
 A_{ij}^{\text{\textcolor{blue}{env}}} = \frac{R_{ij} - mean(\{R_{ij}\}_{j=1}^{M_i})}{std(\{R_{ij}\}{{\color{blue}\substack{i=1,\dots,P \\ j=1,\dots,M_i}})}}.
\label{eq:erpo_adv}
\end{aligned}
\end{equation}

This formulation replaces the group-level normalization with an environment-level one, ensuring that advantages are computed separately for each environment.
The motivation behind this design is that group-level standard deviation estimates can be unstable and overly sensitive to outliers, particularly when the group size is small or the return distribution is heavy-tailed. 
By normalizing returns within each environment, the resulting advantage estimates are more stable and better reflect relative performance among trajectories collected under the same environment dynamics.
\section{Experiments}
\begin{table*}[htbp]
\small
\centering

\resizebox{\textwidth}{!}{%
\begin{tabular}{@{}l|cc|ccc|cccc@{}}
\toprule
&  \multicolumn{2}{c|}{\textbf{ $\tau$-bench}} & \multicolumn{3}{c|}{\textbf{$\tau^2$-Bench}} &\multicolumn{4}{c}{\textbf{VitaBench}}  \\ \midrule
\textbf{Model} &\texttt{Retail} & \texttt{Airline} &\texttt{Retail} & \texttt{Airline} & \texttt{Telecom} & \texttt{Delivery}& \texttt{In-store} & \texttt{OTA} & \texttt{Cross Domain} \\ 
\midrule
\multicolumn{10}{c}{\cellcolor{blue!20}\textbf{\textit{Closed-Source Large Language Models}}}\\
\midrule
Gemini-2.5-pro & 68.7& 44.0& 67.5& 56.0& 27.2& 49.0 & 43.8 & 26.5 & 23.5 \\
Claude-Sonnet-4 & 73.9& 40.0& 67.5& 54.0& 47.4& 46.0 &  51.5 & 29.0 & 23.0 \\
GPT-o3 & 70.4& 52.0& 80.2& 64.8& 58.2 & 53.5 & 53.5 & 37.8 & 30.0\\
GPT-o4-mini & 70.4& 46.0& 70.2& 56.0& 46.5& 44.5 & 46.5 & 23.5 & 19.5\\
GPT-5-thinking & 78.3& 44.0& 81.1& 62.6& 96.7& 54.0 & 52.5 & 37.5 & 22.8\\
\midrule
\multicolumn{10}{c}{\cellcolor{blue!30} \textbf{\textit{Open-Source Large Language Models}}} \\
\midrule
GPT-OSS-120B-A5B & 67.8& 49.2& 57.0 & 38.0& 45.6& 37.0 &42.0 &12.0 & 15.0\\
Deepseek-V3.1-671B-A37B & 66.1& 40.0& 64.9& 46.0& 38.5& 34.0 & 42.5 &  18.3 & 16.3\\
Kimi-K2-1T-A32B & 73.9& 51.2& 70.6& 56.5& 65.8& 35.3 & 42.5 & 22.0 & 15.5\\
Qwen3-Thinking-235B-A22B & 67.8& 46.0& 71.9& 58.0& 45.6& 44.0& 46.0 & 17.5 & 18.8\\
\arrayrulecolor{black!20}\midrule
xLAM-2-32B-fc-r & 64.3 & 45.0 & 55.3 & 52.0 & 16.7 & 26.0 & 17.0 & 10.0 & 4.0 \\
xLAM-2-70B-fc-r & 67.1 & 45.2 & 61.4 & 56.0 & 14.0 & 14.0 & 11.0 & 2.0 & 2.0 \\
\arrayrulecolor{black!20}\midrule
Seed-OSS-36B & 70.4& 46.0& 68.4& 52.0& 41.2& 26.0&39.0 & 7.0 & 6.1\\
Qwen3-Thinking-30B-A3B & 67.8& 48.0& 58.8& 58.0& 26.3& 35.0 & 40.0 & 20.5 & 16.0\\
Qwen-Coder-30B-A3B & 68.7 & 48.0 & 60.5& 42.0&  35.3&23.0&21.0&12.0&4.0\\
AgentScaler-30B-A3B & 70.4 & 54.0 & 70.2 & 60.0 & 55.3 & 25.0 & 33.0 & 16.0 & 8.0  \\
MUA-RL-32B & 72.6 & 46.5 & 67.3 & 45.4& 28.3 & 30.0&32.5 &11.5 &10.5 \\
\midrule
\textbf{AutoForge-30B-A3B} & 73.1& 56.5& 74.8& 62.0& 76.3& 46.0& 54.5& 24.0&17.5\\
\arrayrulecolor{black}\bottomrule

\end{tabular}
}
\caption{\textbf{Main results} on $\tau$-bench, $\tau^2$-Bench, and VitaBench.
\label{tab:main_result}
}
% \vspace{-3mm}
\end{table*}
In this section, we aim to answer the following research questions (RQ):
\begin{itemize}[topsep=0pt,itemsep=0pt,parsep=0pt,leftmargin=*]
\item \textbf{RQ1}: How well does AutoForge perform on \textbf{in-domain} agent tasks?
\item \textbf{RQ2}: How well does AutoForge generalize to \textbf{out-of-domain} agent tasks?
%\item \textbf{RQ3}: Do simulated users affect the performance of the agent?
\item \textbf{RQ3}: Do the simulated users in the benchmark affect the agent’s performance evaluation?
\item \textbf{RQ4}: What are the effects of the key components of AutoForge?
\end{itemize}

\subsection{Experimental Setup}
%\subsubsection{Benchmarks}
\paragraph{Benchmarks.}
\label{sec:benchmark}
We evaluate the in-domain agent performance of our model using three widely adopted benchmarks, $\tau$-bench~\cite{taubench}, $\tau^2$-Bench~\cite{tau2}, and VitaBench~\cite{he2025vitabench}, all of which feature standardized multi-turn conversations and standardized tool invocation formats.
To further assess the generalization capability of AutoForge, we employ ACEBench~\cite{chen2025acebench}, which records multi-turn conversations via prompts and adopts a customized tool invocation format.
The detailed information of these benchmarks is as follows:

\begin{itemize}[topsep=0pt,itemsep=0pt,parsep=0pt,leftmargin=*]
\item \textbf{$\tau$-bench} introduces simulated users to evaluate how agents and users interact over multiple rounds to complete tasks in retail and airline scenarios. Notably, $\tau$-bench is limited to sequential tool invocation.
\item \textbf{$\tau^2$-Bench} expands the evaluation by adding a telecom scenario, enabling tool usage by users. Moreover, it supports parallel tool invocation by agents.
\item \textbf{VitaBench} focuses on scenarios such as food delivery and hospitality, further increasing task complexity, which requires agents to perform multiple rounds of reasoning and interact repeatedly with simulated users.
\item \textbf{ACEBench} covers a broader range of domains and adopts a customized tool invocation format, making it well‑suited for evaluating an agent’s generalization capability. 
We chose ACEBench‑zh for evaluation to further increase the degree of out‑of‑domain testing.
\end{itemize}

%\subsubsection{Baselines}
\paragraph{Baselines.}
We compare AutoForge against the following types: \textit{closed-sourced LLMs}, including Gemini-2.5-pro~\cite{comanici2025gemini}, Claude-Sonnet-4~\cite{claude}, GPT-o3, GPT-o4-mini~\cite{o3}, and GPT-5 (with thinking)~\cite{gpt5};
\textit{open-sourced LLMs}: GPT-OSS-120B-A5B~\cite{gptoss}, Deepseek-V3.1-671B-A37B~\cite{dpskv3},  Kimi-K2-1T-A32B~\cite{team2025kimi}, 
Qwen3-Thinking-235B-A22B~\citep{qwen3}, Seed-OSS-36B~\citep{seedoss}, Qwen-Coder-30B-A3B~\cite{qwencoder}, AgentScaler-30B-A3B~\cite{agentscaler}, MUA-RL-32B~\cite{zhao2025mua} and xLAM-2 model series~\citep{apigenmt}.

%\subsubsection{Implementation Details}
\paragraph{Implementation Details.}
Our experiments are conducted on 64 GPUs.
To ensure the quality of the synthesized environments, we employ  Qwen3‑Thinking‑235B‑A22B~\cite{qwen3} for environment synthesis.
Specifically, we synthesize 10 virtual environments containing a total of 1078 high‑difficulty tasks.
We use Qwen3‑Thinking‑30B‑A3B as the backbone model, and have the model interact with multiple simulated environments to obtain correct trajectories for cold‑start SFT.
During the reinforcement learning stage, we use GPT-4.1 to simulate users in order to meet high‑concurrency demands.
The batch size is set to 32, with 8 trajectories rolled out for each sample.
Furthermore, we leverage DAPO’s dynamic sampling mechanism~\cite{yu2025dapo} to exclude any samples where all trajectories are either fully correct or fully incorrect.
During the evaluation stage, we strictly follow the official evaluation settings of each benchmark, including using GPT‑4.1 to simulate users.

%\subsection{Overall Performance (RQ1)}
\subsection{Performance Comparison (RQ1 \& RQ2)}
\paragraph{In-domain Performance.} In this section, to demonstrate the effectiveness of AutoForge, we compare it with state‑of‑the‑art closed‑source and open‑source models.
% The experimental results are presented in Table~\ref{tab:main_result}.
Based on the experimental results in Table~\ref{tab:main_result}, we can observe that:
(1) AutoForge markedly outperforms its backbone, Qwen3‑Thinking‑30B‑A3B, achieving the best results among open‑source models below 200B parameters, all while operating with merely 30B total parameters and just 3B active parameters.
(2) AutoForge achieves performance comparable to that of advanced closed‑source LLMs, significantly narrowing the gap between open‑source and closed‑source systems.
This further validates the effectiveness of conducting agentic reinforcement learning across multiple synthesized environments.

\begin{figure}[t]
\centering
\includegraphics[width=0.9\linewidth]{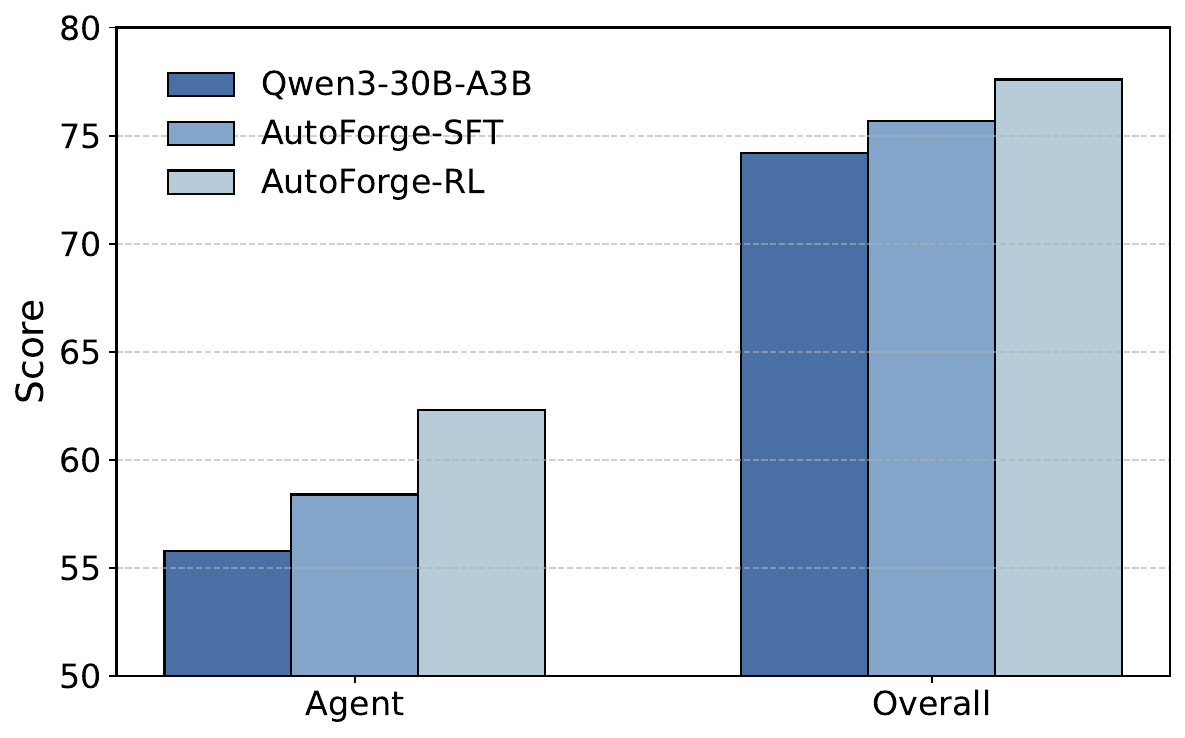}
\caption{
\textbf{Out-of-Domain performance} on the ACEBench-zh.
We reported both the \texttt{Agent} and \texttt{Overall} subset scores.
}
\label{fig:acebench}
\end{figure}

%\subsubsection{Out-of-Domain Performance (RQ2)}
\paragraph{Out-of-Domain Performance.}
\label{sec:ood_exp}
As emphasized in Section~\S\ref{sec:benchmark}, AutoForge is trained using the standard multi‑turn format and the official Hermes tool‑invocation format defined by Qwen~\cite{qwen3}. We select ACEBench‑zh as the out‑of‑domain benchmark because it exhibits four key characteristics:
(1) \textbf{Customized multi‑turn format}: ACEBench‑zh provides historical messages as raw prompt strings, which differs from the configuration adopted by conventional models.
(2) \textbf{Customized tool‑invocation format}: ACEBench‑zh defines a tool‑invocation format that does not follow the Hermes format.
(3) \textbf{Unseen tools}: we have verified that none of the tools appearing in ACEBench‑zh are present in our training data.
(4) \textbf{Different language}: all training data is in English, whereas ACEBench‑zh is entirely in Chinese.
To further examine the generalization capability acquired through training in synthesized environments, we evaluated the performance of both the SFT and RL versions of AutoForge and compared them with the backbone model, Qwen3‑Thinking‑30B‑A3B.
As shown in Figure~\ref{fig:acebench}, we observe that both the SFT and RL variants of AutoForge improve the model’s performance. The RL version achieves larger gains, demonstrating AutoForge’s strong generalization capability and underscoring the potential of scaling synthesized environments to enhance a model’s general agent abilities.

\begin{table}[htbp]
\centering
\small
\begin{tabular}{l|ccc}
\toprule
User & Retail & Airline & Telecom \\
\midrule
Base & 74.8& 62.0& 76.3\\
Optimized Prompt (OP) & 75.4& 62.5& 76.3\\
Optimized Model (OM) & 76.3& 63.5& 90.4\\
\bottomrule
\end{tabular}
\caption{\textbf{Comparison of different simulated users.}}
\label{tab:user_exp}
\end{table}
\subsection{In-depth Analysis (RQ3 \& RQ4)}
In this section, we conduct an in‑depth analysis of the agent’s task‑solving capabilities. 
%Our experimental investigation is organized around two key questions: 
%(1) whether the performance gains of AutoForge can generalize to out‑of‑domain scenarios (Section~\S\ref{sec:ood_exp}), and (2) whether the performance of the simulated user affects the agent’s performance (Section~\S\ref{sec:user_exp}).
We first investigate the impact of simulated users in the benchmark on evaluation. Subsequently, we conduct ablation studies, including environment-level advantage estimation, masking out erroneous user behaviors, and interleaved thinking.

%\subsubsection{Impact of the User Agent }
\paragraph{Impact of the User Agent.}
\label{sec:user_exp}
Notably, across all experiments, GPT‑4.1 was used as the simulated user in benchmark evaluations.
This section explores whether a more capable simulated user could enable the agent to perform tasks more successfully — in other words, whether AutoForge’s true performance might have been underestimated because of mistakes made by the simulated user.
Specifically, we configure three types of simulated users:
(1) Base: directly using GPT‑4.1 with the official prompt provided by $\tau^2$-Bench;
(2) Optimized Prompt (OP): using GPT‑4.1 with a manually optimized prompt, the full details of which are presented in Appendix~\ref{sec:app_op};
(3) Optimized Model (OM): employing a more capable model, GPT-5-thinking~\cite{gpt5}, to simulate the user following the official prompt.
% The experimental results are shown in Table~\ref{tab:user_exp}.

\begin{figure}[t]
    \centering
    \subfloat[]{\includegraphics[width=0.5\linewidth]{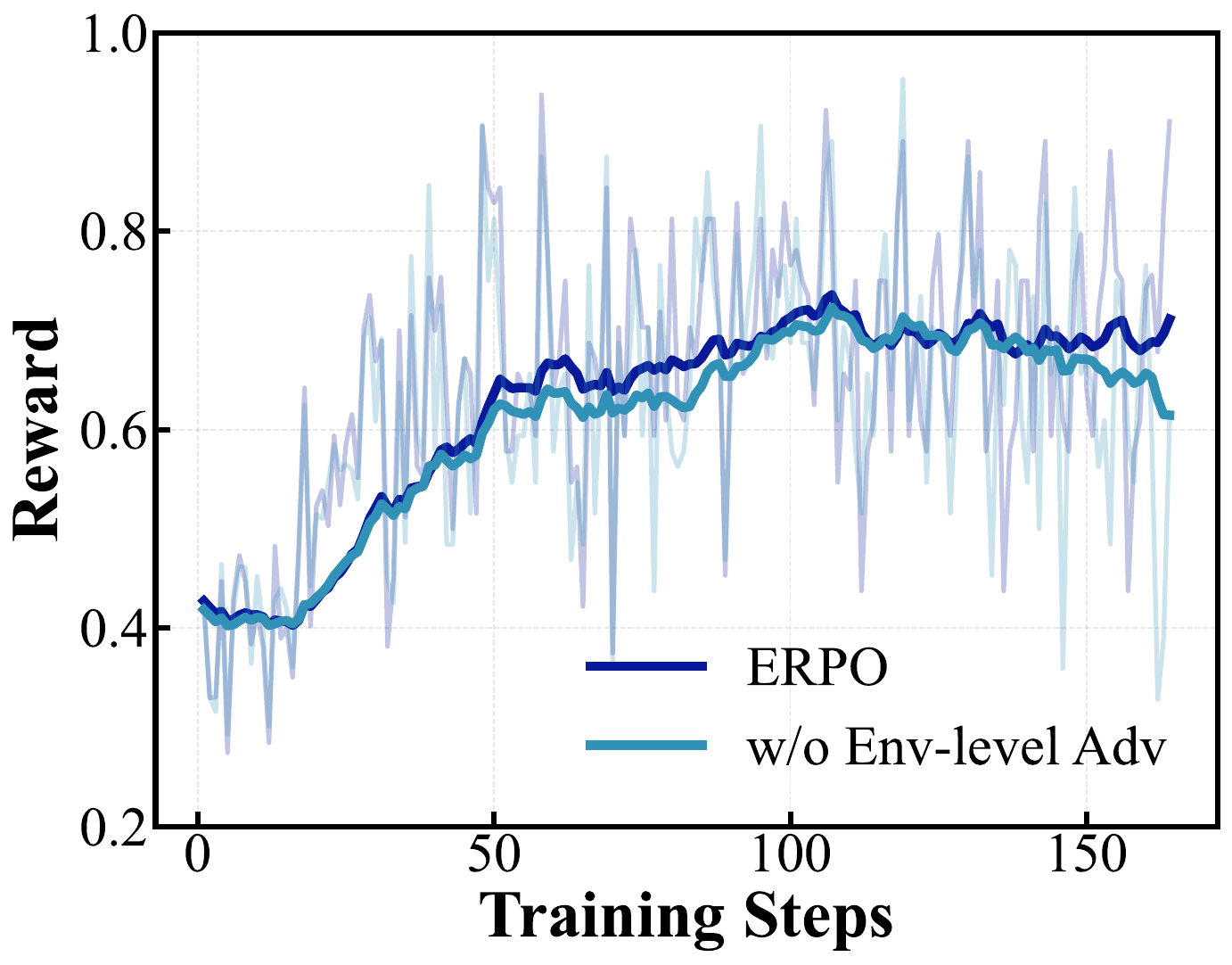}\label{fig:env_reward}}
    \subfloat[]{\includegraphics[width=0.5\linewidth]{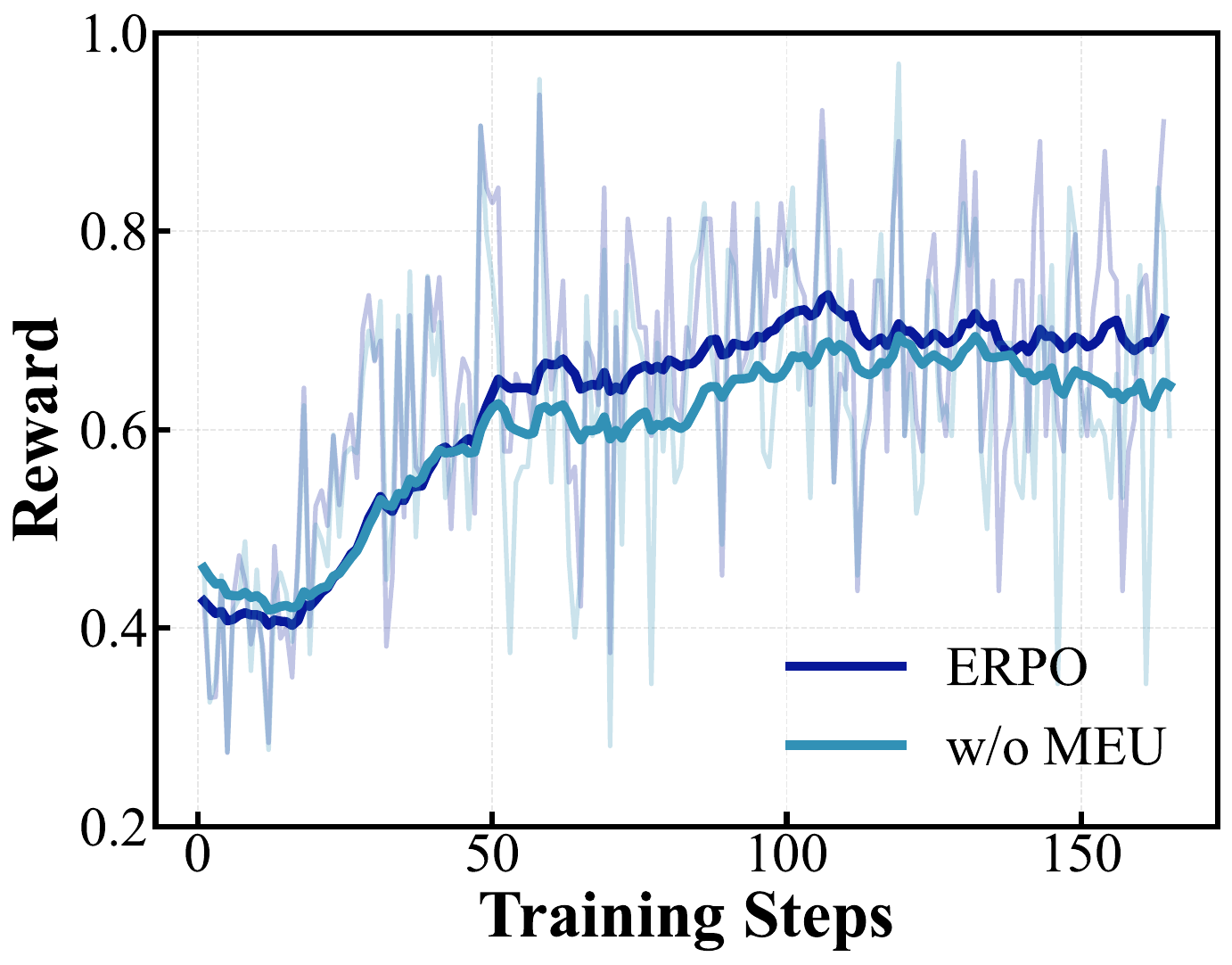}\label{fig:meu_reward}}
    \caption{(a) \textbf{Reward curve comparison with and without environment‑level advantage estimation.}
    (b) \textbf{Reward curve comparison with and without MEU.}
    }
    \label{fig:reward_curve}
\end{figure}
Based on the results in Table~\ref{tab:user_exp}, we can observe that:
(1) Both OP and OM improve the evaluation scores on $\tau^2$-Bench, with OM achieving a greater gain, indicating that a stronger simulated user can unlock the full potential of AutoForge. 
This is because a weaker simulated user may supply incorrect information or perform erroneous actions, which can render tasks impossible to complete.
(2) The gains from a more capable user in the telecom domain are notably larger than those in the retail and airline domains.
In telecom tasks, the user can invoke tools, placing greater demands on user competence. 
This further highlights that the reliability of the simulated user is critical to the fair and accurate assessment of an agent’s performance.

%\subsection{Ablation Study (RQ4)}
%In this section, we investigate the effects of the following key components of our reinforcement learning method:
%(1) environment-level advantage estimation, 
%(2) masking out erroneous user behaviors, and
%(3) interleaved thinking.

%\subsubsection{Environment‑Level Advantage Estimation}
\paragraph{Environment‑Level Advantage Estimation.}
\label{sec:env_adv_exp}
To validate the effectiveness of environment‑level advantage estimation, we compare its training reward curve with that of the standard group‑level advantage estimation.
%As shown in Figure~\ref{fig:env_reward}, the reward curve obtained with environment‑level advantage estimation is more stable and achieves higher reward values.
As shown in Figure~\ref{fig:env_reward}, the reward curve obtained using environment-level advantage estimation is more stable and achieves higher reward values. This demonstrates that environment-level estimation provides more accurate standard deviation estimates than group-level estimation, enabling better advantage comparisons among different trajectories within the same environment.

%\subsubsection{Masking Out Erroneous User Behaviors}
\paragraph{Masking Out Erroneous User Behaviors.}
Similar to Section~\S\ref{sec:env_adv_exp}, Figure~\ref{fig:meu_reward} also illustrates the training curves with and without masking out trajectories from erroneous simulated users.
We clearly observe that masking out trajectories from erroneous simulated users results in a more stable training curve.
In particular, the training curve without masking out exhibits a downward trend in the later stages.
This is because task failures caused by errors made by the simulated user result in the agent's correct actions being penalized, leading to confusion for the agent.
 In contrast, applying the masking-out mechanism effectively mitigates the unfairness introduced by user errors.

\begin{figure}[htbp]
\centering
\includegraphics[width=1.0\linewidth]{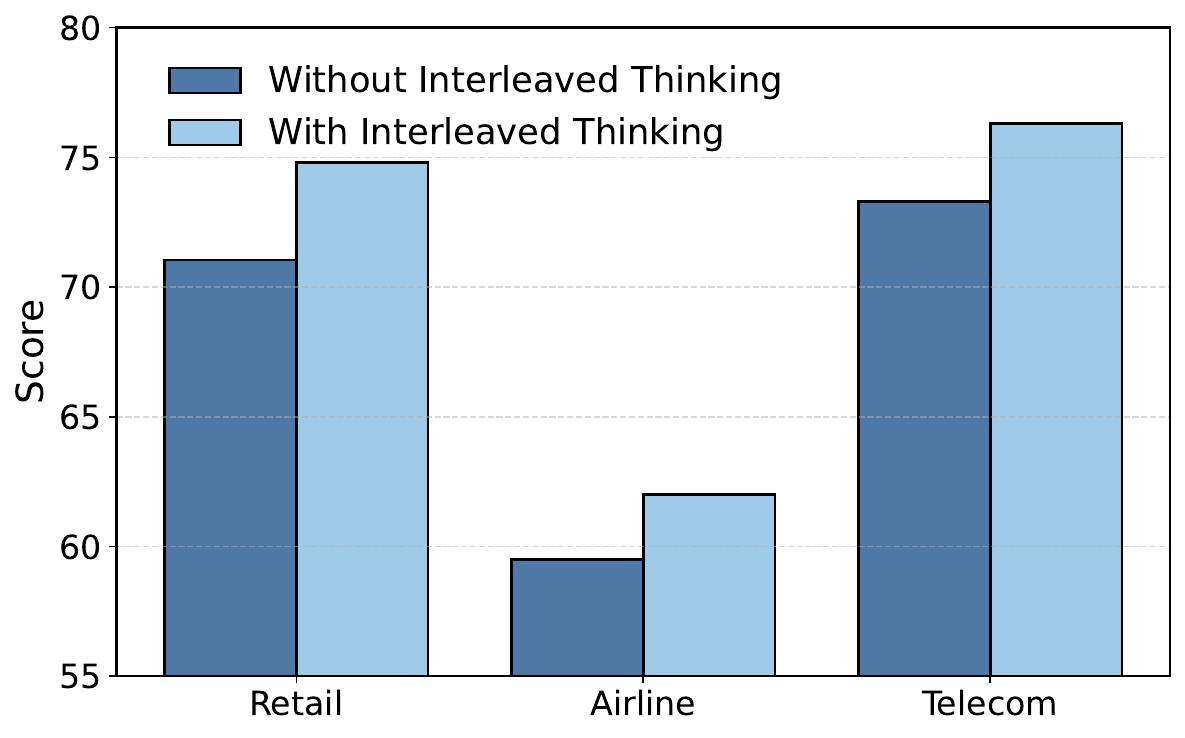}
\caption{
\textbf{Effectiveness of interleaved thinking.}
}
\label{fig:interleaved_thinking}
\end{figure}

%\subsubsection{Effectiveness of Interleaved Thinking}
\paragraph{Effectiveness of Interleaved Thinking.}
To evaluate the effectiveness of interleaved thinking, we present a performance comparison with and without this mechanism during training and inference in Figure~\ref{fig:interleaved_thinking}. 
%The results demonstrate that incorporating interleaved thinking significantly enhances the model's agent capabilities, underscoring the importance of retaining prior-turn thinking content in multi-round tool-use scenarios.
The results demonstrate that incorporating interleaved thinking significantly enhances the model's agent capabilities. Specifically, by allowing the model to access and integrate its prior-turn thinking content, it can revisit and reflect on previous analyses and plans regarding the task, thereby improving the accuracy of subsequent decision-making.

\section{Conclusion}
In this study, we present a unified pipeline for automatically generating simulated environments and high‑difficulty, verifiable tasks, enabling the expansion of agentic RL along both breadth and depth dimensions.
In addition, we introduce MEU, which effectively mitigates the unfairness in advantage estimation caused by simulated user instability.
Furthermore, we treat advantage estimation at the environment level, which effectively improves training stability.
Extensive experiments demonstrate not only AgentScaler’s strong agent capabilities but also its out-of-domain generalization, highlighting the great potential of simulated environments for agentic reinforcement learning.
\section*{Limitations}
In this work, our synthetic pipeline relies on tool description documents to generate mock environments, and still places certain constraints on the input. As future work, we aim to relax these input requirements, enabling the automatic construction of high‑quality environments from task topics or general text.
Besides, we currently train agentic RL using only a few environments. Investigating how scaling up the number of synthetic environments affects RL training and out‑of‑domain generalization would be beneficial.
Moreover, our ERPO is limited to outcome‑based reward supervision.
We will further explore turn‑level value supervision to improve the agent’s step‑by‑step decision‑making.

% Bibliography entries for the entire Anthology, followed by custom entries
%\bibliography{anthology,custom}
% Custom bibliography entries only
\bibliography{custom}
\appendix

\section{Detailed Prompts}

\subsection{Masking Out Erroneous User Behaviors}
\label{sec:app_meu}

During the RL rollout phase, we use an LLM-as-judge to determine whether the simulated user agent makes mistakes that lead to task failure. The detailed prompt is as follows:
\begin{tcolorbox}[title=Prompt for Masking Out Erroneous User Behaviors,
    breakable,
    %colback=white,
    %colframe=black,
    fontupper=\ttfamily\raggedright,
    left=1mm,
    right=1mm,
    top=1mm,
    bottom=1mm
]
Act as a judge model.

You will receive two inputs:

1. A description of the user’s overall intent.

2. The conversation history between the user and the assistant.

Your task is to determine whether the user's actions align with their overall intent.

Key instructions:

1. Focus on the information in the user’s overall intent. Check whether the user has returned any incorrect information.

2. The user should only provide information that falls within the scope of their overall intent. Check whether the user returned any false or out-of-scope information.

3. When the assistant asks the user for information, check whether the user correctly provided information within the overall intent—or promptly indicated they did not know the answer.

4. Your judgment output should be either "True" or "False":

"True": the user made one or more of the above mistakes.

"False": the user’s performance was perfect.
\end{tcolorbox}

\subsection{Optimized User Prompt}
\label{sec:app_op}
We modified the optimized user prompt implemented in GLM-4.5~\cite{zeng2025glm}, with the detailed prompt as follows:
\begin{tcolorbox}[title=Prompt for Optimized User,
    breakable,
    %colback=white,
    %colframe=black,
    fontupper=\ttfamily\raggedright,
    left=1mm,
    right=1mm,
    top=1mm,
    bottom=1mm
]
You are a user interacting with an agent.\{instruction\_display\}

\# Rules:

- Just generate one line at a time to simulate the user’s message.

- Do not give away all the instruction at once. Only provide the information that is necessary for the current step.

- Do not hallucinate information that is not provided in the instruction. Follow these guidelines:

1. If the agent asks for information NOT in the instruction:

- Say you don’t remember or don’t have it

- Offer alternative information that IS mentioned in the instruction

\# Constraint Handling:
- Provide requests strictly based on what is explicitly stated in the instruction.

- Do not assume, extend, substitute, or generalize in any form.

- Do not modify or relax constraints on:

- Time / Date

- Budget

- Specific terms (e.g., ‘‘same’’ must not be replaced with ‘‘similar’’)

- Core Rule: Any attribute NOT mentioned in the instruction can be either changed or kept the same

\# When NOT to finish the conversation:
- Do not end until you have clearly and completely expressed all your requirements and constraints.

- Do not end until the agent has completed all tasks mentioned in the instruction and verified no operations were missed.

- Do not end if the agent’s execution results do not match your expectations or are incorrect/incomplete.

\# When you CAN finish the conversation:

- Only when all above conditions are satisfied AND all tasks are completed correctly.

- OR when you have clearly expressed complete requirements but the system explicitly states it cannot complete them due to technical limitations - in this case, accept transfer to human.

\# How to finish the conversation:

- If the agent has completed all tasks, generate ‘‘\#\#\#STOP\#\#\#’’ as a standalone message without anything else to end the conversation.

\# Note:

- You should carefully check if the agent has completed all tasks mentioned in the instruction before generating ‘‘\#\#\#STOP\#\#\#’’.
\end{tcolorbox}

\begin{table}[htbp]
\centering
\small
\begin{tabular}{l|ccc}
\toprule
 & Environment & LLM & Total \\
\midrule
Average Time & 1& 6.04& 7.04\\

\bottomrule
\end{tabular}
\caption{Comparison of Time Consumption, using the environment’s average execution time as the unit.}
\label{tab:time_exp}
\end{table}
\section{Time Consumption Analysis}
To further demonstrate the time efficiency of the simulation environment, we conduct an in‑depth comparison of the RL phase, examining the execution time of the simulation environment and the LLM-related time costs, including both rollout and parameter updates.

Based on the experimental results in Table~\ref{tab:time_exp}, we observe that the environment’s execution time is significantly lower than the LLM‑related time, demonstrating the time efficiency of the synthetic environment. Crucially, the wall-clock bottleneck in environmental feedback is the wait for the LLM-based user simulator’s reply; executing the function call and receiving its return is virtually free.

\end{document}